\documentclass[10pt,twocolumn,letterpaper]{article}

\usepackage{cvpr}
\usepackage{times}
\usepackage{epsfig}
\usepackage{graphicx}
\usepackage{amsmath}
\usepackage{amssymb}
\usepackage{capt-of}
\usepackage{verbatim}
\usepackage{mathrsfs}

\usepackage[linesnumbered,lined,ruled]{algorithm2e}

\usepackage[breaklinks=true,bookmarks=false]{hyperref}

\cvprfinalcopy 

\def\cvprPaperID{****} 

\begin{document}

\title{Joint Neural Architecture Search and Quantization}

\author{
Yukang Chen$^{1,2}$, Gaofeng Meng$^{1,2}$, Qian Zhang$^3$ \\ Xinbang Zhang$^{1,2}$, Liangchen Song$^3$, Shiming Xiang$^{1,2}$, Chunhong Pan$^{1,2}$\\
$^1$\emph{National Laboratory of Pattern Recognition, Institute of Automation, Chinese Academy of Sciences}\\
$^2$\emph{University of Chinese Academy of Sciences} $\; ^3$\emph{Horizon Robotics} \\
\emph{\{yukang.chen, gfmeng, xinbang.zhang, smxiang, chpan\}@nlpr.ia.ac.cn}\\ 
\emph{\{qian01.zhang, liangchen.song\}@horizon.ai}
}

\maketitle
\begin{abstract}

Designing neural architectures is a fundamental step in deep learning applications. As a partner technique, model compression on neural networks has been widely investigated to gear the needs that the deep learning algorithms could be run with the limited computation resources on mobile devices. Currently, both the tasks of architecture design and model compression require expertise tricks and tedious trials.
In this paper, we integrate these two tasks into one unified framework, which enables the joint architecture search with quantization (compression) policies for neural networks. This method is named JASQ.
Here our goal is to automatically find a compact neural network model with high performance that is suitable for mobile devices. Technically, a multi-objective evolutionary search algorithm is introduced to search the models under the balance between model size and performance accuracy.

In experiments, we find that our approach outperforms the methods that search only for architectures or only for quantization policies. 1) Specifically, given existing networks, our approach can provide them with learning-based quantization policies, and outperforms their 2 bits, 4 bits, 8 bits, and 16 bits counterparts. It can yield higher accuracies than the float models, for example, over 1.02\% higher accuracy on MobileNet-v1.
2) What is more, under the balance between model size and performance accuracy, two models are obtained with joint search of architectures and quantization policies: a high-accuracy model and a small model, JASQNet and JASQNet-Small that achieves 2.97\% error rate with 0.9 MB on CIFAR-10.

\end{abstract}

\section{Introduction}
\begin{figure*}[htb]
  \centering
  \includegraphics[width=1.0\textwidth]{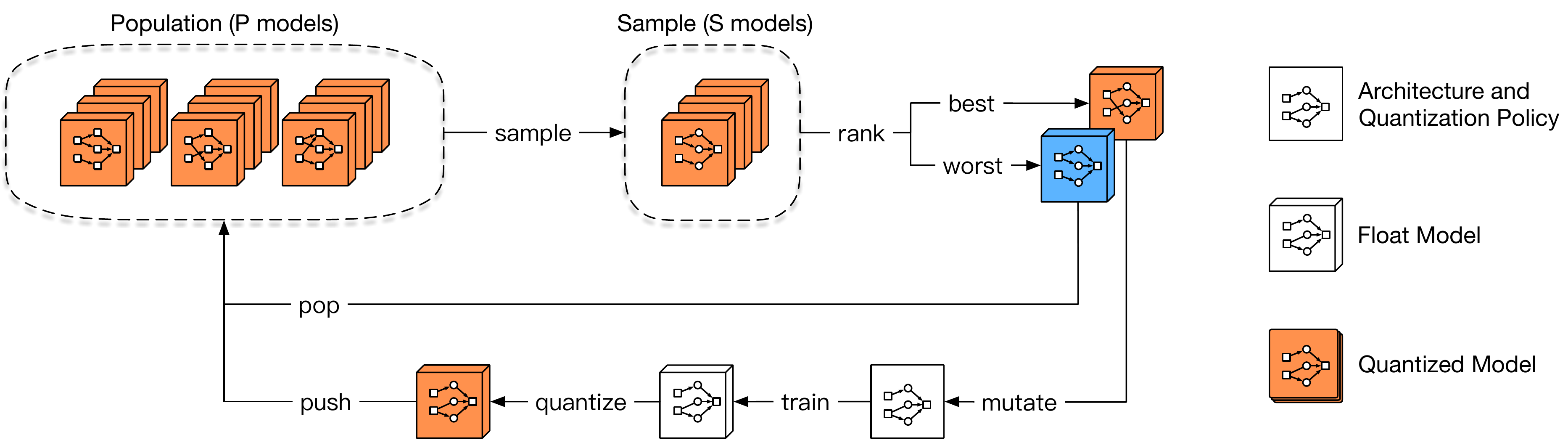}
  \caption{The evolutionary algorithm framework for our joint search method. Each individual in the population is evaluated with the accuracy and
model size of the quantized model.  
  When architectures are fixed during search, the method could provide existing networks with quantization policies.}\label{fig:1}
\end{figure*}
Deep convolutional neural networks have successfully revolutionized various challenging tasks, e.g., image classification~\cite{he2016deep,huang2017densely,szegedy2016rethinking}, object detection~\cite{DBLP:journals/pami/RenHG017} and semantic segmentation~\cite{DBLP:journals/pami/ChenPKMY18}.
Benefited from its great representation power, CNNs have released human experts from laborious feature engineering with end-to-end learning paradigms.
However, another exhausting task appears, i.e., neural architecture design that also requires endless trails and errors. 
For further liberation of human labours, many neural architecture search (NAS) methods~\cite{zoph2017learning,Real2018Regularized} have been proposed and proven to be capable of yielding high-performance models. But the technique of NAS alone is far from real-world AI applications.  

As networks usually need to be deployed on devices with limited resources, model compression techniques are also indispensable. 
In contrast to NAS that is considered at the topological level, model compression aims to refine the neural nodes of a given network with sparse connections or weighting-parameter quantization.
 However, computation strategies also need elaborate design. Taking quantization for example, conventional quantization policies often compress all layers to the same level. 
Actually each layer has different redundancy, it is wise to determine a suitable quantization bit for each layer.
However, quantization choices also involve a large search space and designing mutual heuristics would make human burden heavier. 

In this paper, we make a further step for the liberation of human labours and propose to integrate architecture search and quantization policy into a unified framework for neural networks (JASQ).
A Pareto optimal model~\cite{deb2014multi} is constructed in the evolutionary algorithm to achieve good trade-offs between accuracy and model size.
By adjusting the multi-objective function, our search strategy can output suitable~models for different accuracy or model size demands.
During search, a population of models are first initialized and then evolved in iterations according to their fitness. Fig.~\ref{fig:1} shows the evolutionary framework of our method. 
Our method brings the following advantages:
\begin{itemize}
\item \emph{Effectiveness} $\;$ Our method can jointly search for neural architectures and quantization policies. The resulting models, i.e., JASQNet and JASQNet-Small, achieve competitive accuracy to state-of-the-art methods~\cite{he2016deep,huang2017densely,zoph2017learning} and have relatively small model size.
For existing architectures, e.g., ResNet~\cite{he2016deep}, DenseNet~\cite{huang2017densely} and MobileNets~\cite{howard2017mobilenets,sandler2018inverted}, our quantized models can outperform their 2/4/8/16 bits counterparts and even achieve higher accuracies than float models on ImageNet.

\item \emph{Flexibility} $\;$ 
 In our evolutionary search method, a multi-objective function is adopted as illustrated in Fig.~\ref{fig:3} and Eq.~\eqref{Eq1}. By adjusting $\mathcal{T_S}$ in the objective function, we obtain models with different accuracy and size balances. JASQNet has a comparable accuracy to ResNet34~\cite{he2016deep} but much less model size.  JASQNet-Small has a similar model size to SqueezeNet~\cite{DBLP:journals/corr/IandolaMAHDK16} but much better accuracy (65.90\% vs 58.09\%).

\item \emph{Efficiency} $\;$ 
We need only 1 GPU across 3 days to accomplish the joint search of architectures and quantization policies.
Given hand-craft networks, their quantization policies can be automatically found in a few hours on ImageNet.

\end{itemize}

\begin{figure*}[t]
  \centering
  \includegraphics[width=1.0\textwidth]{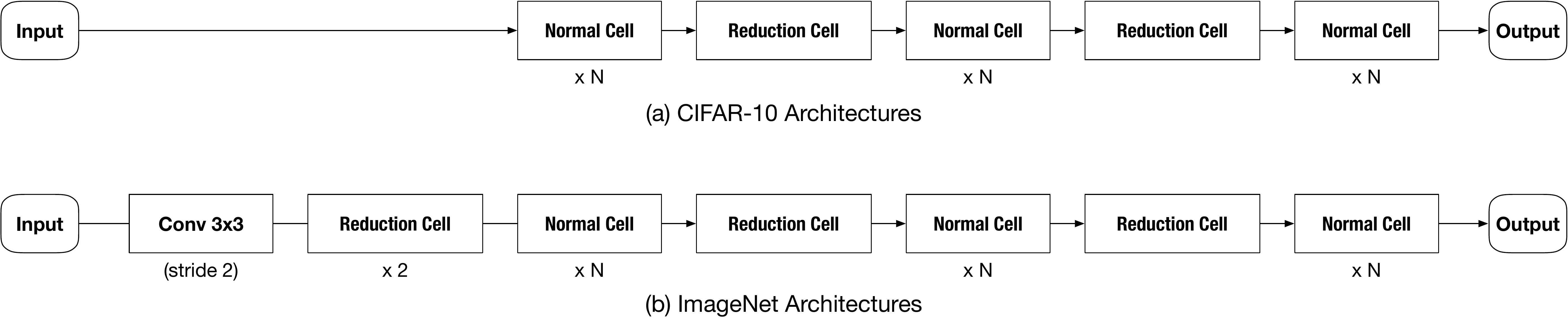}
  \caption{Architectures for CIFAR-10 and ImageNet. 
  The image size in ImageNet (224x224) is much larger than that in CIFAR-10 (32x32). So there are additional reduction cells and convolution 3x3 with stride 2 in ImageNet architectures to downsample feature maps.}\label{fig:2}
\end{figure*}

\section{Related Work}

\subsection{Neural Architecture Search}
Techniques in automatically designing network~\cite{zoph2017learning,pham2018efficient,Real2018Regularized} have attracted increasing research interests. Current works usually fall into one of two categories: reinforcement learning (RL) and evolutionary algorithm (EA). 
In terms of RL-based methods, NAS~\cite{zoph2016neural} abstracts networks into variable-length strings and uses a reinforcement controller to determine models sequentially. NASNet~\cite{zoph2017learning} follows this search algorithm, but adopts cell-wise search space to save computational resources. 
In terms of EA-based methods, AmoebaNet~\cite{Real2018Regularized} shows that a common evolutionary algorithm without any controller can also achieve comparable results and even surpass RL-based methods. 

In addition to RL and EA, some other methods have also been applied.
DARTS~\cite{DBLP:journals/corr/abs-1806-09055} introduces a gradient-based method where they formulate the originally discrete search space into continuous parameters.
PNAS~\cite{liu2017progressive} uses a sequential model-based optimization (SMBO) strategy to search architectures in order of increasing complexity.
Other methods including MCTS~\cite{DBLP:journals/corr/NegrinhoG17}, boosting~\cite{DBLP:conf/icml/CortesGKMY17} and hill-climbing~\cite{DBLP:journals/corr/abs-1711-04528} have also shown their potentials.
Most methods mentioned above have produced networks that outperforms classical hand-crafted models.
However, only neural architectures can not satisfy the demands of real-world applications. Thus, we propose a more convenient approach to provide complete schemes for deep learning practitioners.

\subsection{Model Compression}   
Model compression has received increasing attention. This technique can effectively execute deep models in resource-constrained environments, such as mobile or embedded devices. A few practical methods are proposed and put into effect. Network pruning conducts channel-level compressions for CNN models~\cite{DBLP:conf/iccv/LuoWL17, DBLP:journals/corr/HanMD15}. Distillation has been introduced recently~\cite{DBLP:journals/corr/HintonVD15, DBLP:conf/nips/BaC14} that transfers the behaviour of a given model to the smaller student structure. In addition, some special convolution structures are also applied in mobile size devices, such as separable depthwise convolution~\cite{howard2017mobilenets} and 1 x 3 then 3 x 1 factorized convolution~\cite{szegedy2016rethinking}. To reduce the redundancy of the fully connected layer, some methods propose to factorize its weights into truncated pieces~\cite{DBLP:conf/nips/DentonZBLF14,DBLP:conf/interspeech/XueLG13}.

Quantization is also a significant branch of model compression and widely used in real applications~\cite{DBLP:journals/corr/abs-1802-05668,DBLP:journals/corr/ZhuHMD16,DBLP:conf/eccv/RastegariORF16}. Quantization can effectively reduce model size and thus save storage space and communication cost. 
Previous works tend to use a uniform precision for the whole network regardless of the different redundancy for each layer. Determining mixed precisions for different layers seems more promising.
Actually mixed precision storage and computation have been widely supported by most hardware platforms, e.g., CPUs and FPGAs. However, because each model has tens or hundreds of layers, it is tedious to conduct this job by human experts. In this work, we combine the search of quantization policies with neural architecture search. Determining a quantization bit for a convolution layer is similar to choosing its kernel size. It is easy to implement this method based on previous NAS works.

\section{Methods}
Neural architecture design and model compression are both essential steps in deep learning applications, especially when we face mobile devices that have limited computation resources. However, both of them are time-consuming if conducted by human experts. 
In this work, we joint search of neural architectures and quantization policies in a unified framework. Compared with only searching for architectures, we evolve both architectures and quantization policies and use the validation accuracies of quantized models as fitnesses. Fig.~\ref{fig:1} illustrates our framework. 

\begin{figure}
  \centering
  \includegraphics[width=0.5\textwidth]{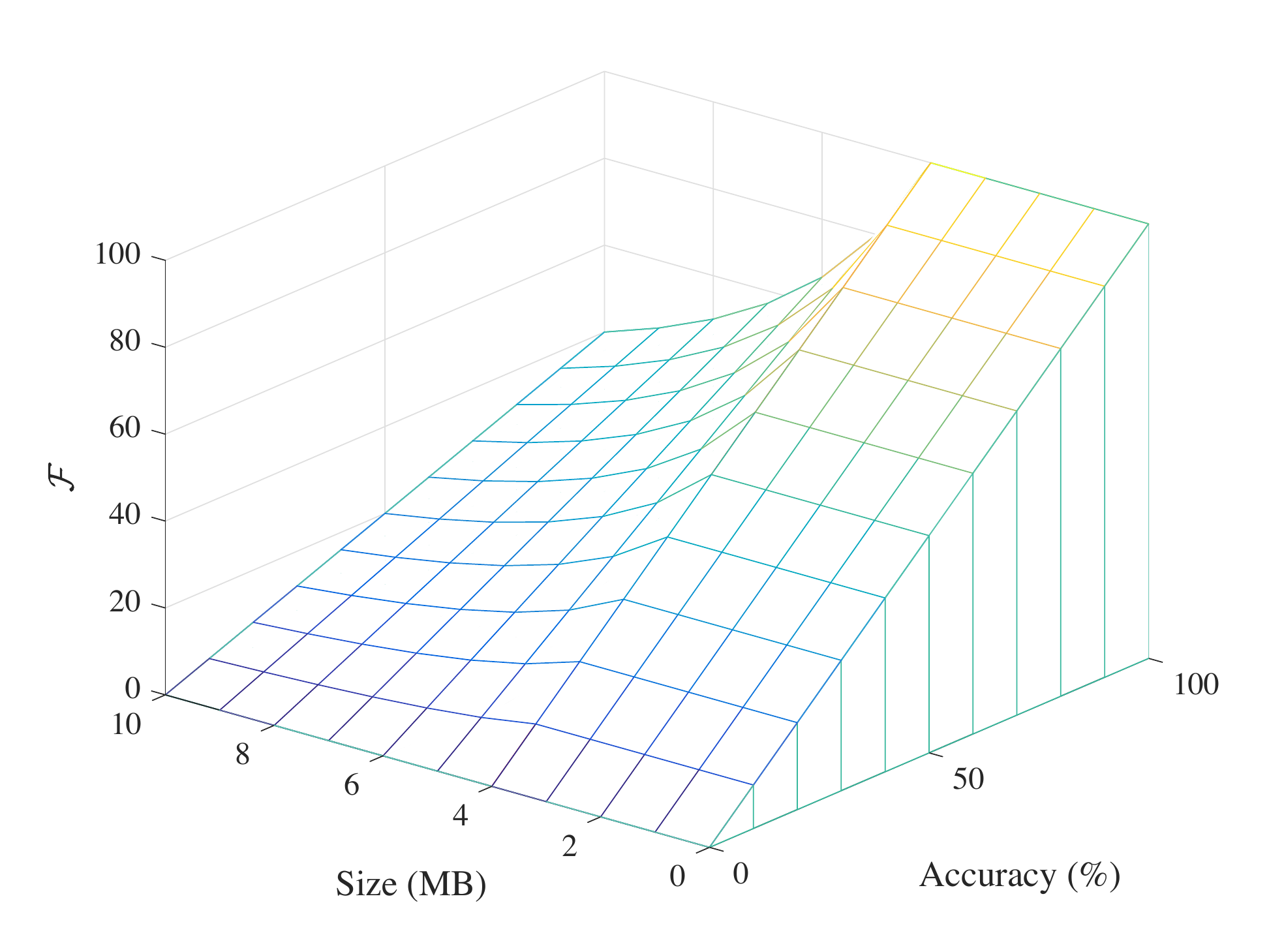}
  \caption{Multi-objective Function. Take $\mathcal{T_{S}}=4~\mathtt{MB}$ for example. When size is less than $\mathcal{T_{S}}$, $\mathcal{F}$ depends only on accuracy. Otherwise, $\mathcal{F}$ sharply decreases as punishment.}\label{fig:3}
\end{figure}
\subsection{Problem Definition}
\label{3.1}
A quantized model $\Theta$ can be constructed by its neural network architecture $\mathscr{A}$ and its quantization policy $\mathscr{P}$.
After the model is quantized, we can obtain its validation accuracy $\mathcal{\alpha}$($\Theta$) and its model size $\mathcal{S}(\Theta)$.
In this paper, we define the search problem as a multi-objective problem. 
The Pareto optimal model~\cite{deb2014multi} is famous for solving multi-objective problems and we define our search problem into maximizing the objective function $\mathcal{F}(\Theta)$ as follow:
 \\
\begin{equation}\label{Eq1}
\begin{split} 
\underset{\small{\Theta}} {\text{max}}
\; \mathcal{F}(\Theta)=
\underset{\small{\Theta}} {\text{max}}
\; \mathcal{\alpha}(\Theta) \cdot \left[\frac{\mathcal{S}(\Theta)}{\mathcal{T_{S}}}\right]^\gamma
\end{split}
\end{equation}
where $\mathcal{T_{S}}$ is the target for the model size and $\gamma$ in the formulation above is defined as follow:

\begin{equation}\label{Eq2}
\begin{split}
 \gamma=\left\{
\begin{aligned}
& 0, \;\;\;\; \text{if}\;\; \mathcal{S}(\Theta) \leq \mathcal{T_{S}} \\
& -1,   \;\;\;\; \text{otherwise}\\
\end{aligned}
\right.
\end{split}
\end{equation}
 It means that if the model size meets the target, we simply use accuracy as the objective function. It degrades to a single objective problem. Otherwise, the objective value is penalized sharply to discourage the excessive model size. We visualize the multi-objective function in Fig.~\ref{fig:3}.

The search task is converted into finding a neural architecture $\mathscr{A}$ and a quantization policy $\mathscr{P}$ to construct an optimal model $\Theta=\{\mathscr{A},\mathscr{P}\}$
that maximizes the objective Eq.~\eqref{Eq1}. 
In experiments, we first show the effectiveness of the learned quantization policies by fixing the network architecture $\mathscr{A}$ as classical hand-crafted networks. After that, the whole search space is explored as described in Section~\ref{3.2}.

\subsection{Search Space}
\label{3.2}
Our search space can be partitioned into neural architecture search space and quantization search space, $\mathbb{S}=\{\mathbb{S}_{\mathscr{A}},\mathbb{S}_{\mathscr{P}}\}$. In this section, we first introduce them respectively and then summarize our total search space in details.

For neural architecture search space $\mathbb{S}_{\mathscr{A}}$, we follow the NASNet search space~\cite{zoph2017learning}.
This search space has been widely used by many well-known methods~\cite{pham2018efficient,Real2018Regularized,liu2017progressive,DBLP:journals/corr/abs-1806-09055} and thus it is fair for comparison. This cell-wise search space consists of two kinds of Inception-like modules, called the \emph{normal cells} and the \emph{reduction cells}. When taking in a feature map as input, the \emph{normal cells} return a feature map of the same dimension. The \emph{reduction cells} return a feature map with its height and width reduced by a factor of two. These cells are stacked in certain patterns for CIFAR-10 and ImageNet respectively as shown in Fig.~\ref{fig:2}. The resulting architecture is determined by the normal cell structure and the reduction cell structure, the first convolution channels ${(F)}$ and cell stacking number ${(N)}$. Only the structure of the cells are altered during search. Each cell is a directed acyclic graph consisting of combinations. A single combination takes two inputs and applies an operation to each of them.
Therefore, each combination can be specified by two inputs and two operations,  $\{{i_1,i_2,o_1,o_2}\}$. 
 The combination output is the addition of them and all combination outputs are concatenated as the cell output.

For quantization policy $\mathbb{S}_{\mathscr{P}}$, we aim to find optimal quantization bit for each cell. As shown in Fig.~\ref{fig:2}, there are ${k = 3\cdot N+2}$ cells in the CIFAR-10 architecture. Thus, the problem is convert into searching for a string of bits for these cells $\mathscr{P} =\{{b_1,b_2,...,b_{k}}\}$. 

In our implementation, we conduct search with a string of code to represent our total search space $\mathbb{S}$. As the neural architecture is determined by the normal cell and the reduction cell, each model is specified by the normal cell structure and the reduction cell structure, $\mathbb{S}_{\mathscr{A}}=\{\mathscr{A}_{\mathtt{nom}},\mathscr{A}_{\mathtt{rec}}\}$. As mentioned above, the normal cell structure contains ${k=3\cdot N+2}$ combinations, that is, $\mathscr{A}_{\mathtt{nom}} = {\{C_1,C_2,...,C_k\}_\mathtt{nom}}$ and the reduction cell structure is same. A combination is specified by two inputs and two operations, which is presented as ${C_j=\{i_1,i_2,o_1,o_2\}_j}$.
The choices of architecture operations ${o}$ and quantization levels ${b}$ are shown below:

 \emph{$\bullet$ Architecture}: 3x3 separable conv, 5x5 separable conv, 3x3 avg pooling, 3x3 max pooling, zero, identity.
 
 \emph{$\bullet$ Quantization}: 4 bit, 8 bit, 16 bit.

Assuming there are $\text{\#}\mathbb{S}_{\mathscr{A}}$ possible architectures and $\mathtt{\text{\#}}\mathbb{S}_{\mathscr{P}}$ possible compression heuristics respectively, the total complexity of our search space is $\mathtt{\text{\#}\mathbb{S}_{\mathscr{A}} \cdot \text{\#}\mathbb{S}_{\mathscr{P}}}$. In experiments, we search on CIFAR-10 and the cell stacking number $\mathtt{(N)}$ is 6. As in Fig.~\ref{fig:2}, there are ${6\times 3 + 2=20}$ cells in each model and $\mathtt{\text{\#}}\mathbb{S}_{\mathscr{P}}$ equals to ${3^{20}=3.5\times 10^9}$. For architecture search space, all our comparison methods and our approach follow. NASNet~\cite{zoph2017learning}.
Thus, our total search space is ${3.5\times 10^9}$ times large as that of comparison methods.

\subsection{Search Strategy}
\label{3.3}
We employ a classical evolutionary algorithm, \emph{tournament selection}~\cite{goldberg1991comparative}.
A population of models $\mathbf{P}$ is first initialized randomly. For any model $\Theta$, we need to optimize its architecture $\mathscr{A}$ and quantization policy $\mathscr{P}$.
 Each individual model $\Theta$ of $\mathbf{P}$ is first trained on the training set $D_{train}$, quantized as its compression strategy and then evaluated on the validation set $D_{val}$. Combined with its model size $\mathcal{S}(\Theta)$, its fitness  $\mathcal{F}(\Theta)$ is computed as Eq.~\eqref{Eq1}. At each evolutionary step, a subset $\mathbf{S}$ is randomly sampled from $\mathbf{P}$.
 According to their fitnesses, we can select the best individual $\Theta_{\mathtt{best}}$ and the worst individual $\Theta_{\mathtt{worst}}$ among $\mathbf{S}$.
 $\Theta_{\mathtt{worst}}$ is then excluded from $\mathbf{P}$ and $\Theta_{\mathtt{best}}$ becomes a parent  and produces a child $\Theta_{\mathtt{mut}}$ with \emph{mutation}.
   $\Theta_{\mathtt{mut}}$ is then trained, quantized and evaluated to measure its fitness $\mathcal{F}(\Theta)$. Afterwards $\Theta_{\mathtt{mut}}$ is pushed into $\mathbf{P}$. 
   This scheme actually keeps repeating competitions of random samples in iterations. The procedure is formulated in Algorithm \ref{algo:1}.

 Specially, \emph{mutation} is conducted to the neural architecture $\mathscr{A}$ and the quantization policy $\mathscr{P}$ respectively in each iteration. For neural architecture $\mathscr{A}$, we make mutations to each combination in the cells, that is to randomly choose one from $\{{i_1,i_2,o_1,o_2}\}$, and then replace it with a random substitute. For quantization policy $\mathscr{P}$, \emph{mutation} is to randomly pick one from $\{{b_1,b_2,...,b_{k}}\}$ and reset it as a random choice of quantization bits.

\begin{algorithm}[t]
\caption{Search Strategy} \label{algo:1}
\SetKwInOut{Input}{input}\SetKwInOut{Output}{output}
\Input{population size $\mathtt{\text{\#}P}$, sample size $\mathtt{\text{\#}S}$, \\
training set \texttt{$D_{train}$}, validation set \texttt{$D_{val}$}, \\
max num epochs $\mathtt{\text{\#}E}$ }

\Output{a population of models $\mathbf{P}$}
\begin{small}
\texttt{$\mathbf{P^{(0)}}$ $\gets$ initialize($\mathtt{\text{\#}E}$)}


\For{i=1:$\mathtt{\text{\#}E}$}{

		\texttt{$\mathbf{S^{(i)}}$ $\gets$ sample($\mathbf{P^{(i-1)}}$, $\mathtt{\text{\#}S}$)}
		
		\texttt{$\Theta_{\mathtt{best}}$,$\Theta_{\mathtt{worst}}$ $\gets$ select($\mathbf{S^{(i)}}$)}
		
		\texttt{$\mathscr{A}_{\mathtt{mut}}$ $\gets$ mutate($\mathscr{A}_{\mathtt{best}}$)}
		
		\texttt{$\mathscr{P}_{\mathtt{mut}}$ $\gets$ mutate($\mathscr{P}_{\mathtt{best}}$)}

		\texttt{$\Theta_{\mathtt{mut}}$ $\gets$ train($D_{train}$, $\mathscr{A}_{\mathtt{mut}}$)}
				
		\texttt{$\mathcal{S}(\Theta_{\mathtt{mut}})$ $\gets$ quantize($\Theta_{\mathtt{mut}},\mathscr{P}_{\mathtt{mut}}$)}

		\texttt{$\mathcal{\alpha}(\Theta_{\mathtt{mut}})$  $\gets$ test($\Theta_{\mathtt{mut}}$, $D_{val}$)}

		\texttt{$\mathcal{F}(\Theta_{\mathtt{mut}})$ $\gets$ Eq.\eqref{Eq1}($\mathscr{\alpha}(\Theta_{\mathtt{mut}})$ ,$\mathcal{S}(\Theta_{\mathtt{mut}})$)}

		\texttt{$\mathbf{P^{(i-1)}}$ $\gets$ push($\mathbf{P^{(i-1)}}$, $\Theta_{\mathtt{mut}}$)}
		
		\texttt{$\mathbf{P^{(i)}}$ $\gets$ pop($\mathbf{P^{(i-1)}}$, $\Theta_{\mathtt{worst}}$)}

}
\end{small}
\end{algorithm}

\begin{table*}[ht]
		\small
		\centering
  \captionof{table}{The results of quantization policy search for existing networks on ImageNet. Here we compare to 8 bits models and float models.
  Numbers in brackets are Acc increase and Size compression ratio compared to float models.}
 \label{tab:1}

  \begin{tabular}{ l | c r | c r | c c }
      \hline
   			& \multicolumn{2}{c|}{\textbf{Ours}} 		   & \multicolumn{2}{c|}{\textbf{8 bits}} 					& \multicolumn{2}{c}{\textbf{Float}}\\ 
            & Acc/\% & Size/MB & Acc/\% & Size/MB & Acc/\% & Size/MB \\
      \hline

	ResNet18~\cite{he2016deep}			 		& \textbf{70.02 (+0.26)}&\textbf{7.21 (6.49x)}		   & 69.64 (-0.12)&11.47 (4.08x)	    			& 69.76&46.76\\
	ResNet34~\cite{he2016deep} 					& \textbf{73.77 (+0.46)}&\textbf{11.92 (7.31x)}		   & 73.23 (-0.08)&21.32 (4.09x)	   				&73.31&87.19\\
	ResNet50~\cite{he2016deep} 					& \textbf{76.39 (+0.26)}&\textbf{14.91 (6.86x)}		   & 76.15 (+0.02)&24.74 (4.13x)	   				&76.13&102.23\\ 
	ResNet101~\cite{he2016deep} 					& 	\textbf{78.13 (+0.76)}&\textbf{31.54 (5.65x)}	 	   & 77.27 (-0.10)&43.19 (4.12x)	   				&77.37&178.20\\ 
	ResNet152~\cite{he2016deep} 					& 	\textbf{78.86 (+0.55)}&\textbf{46.63 (5.16x)}	   	   & 78.30 (-0.01)&58.38 (4.12x)	   				&78.31&240.77\\ 
      \hline
	DenseNet-121~\cite{huang2017densely} 			& \textbf{74.56 (+0.12)}&\textbf{6.15 (5.19x)}	       & 74.44 (+0.00)&7.65 (4.17x)					&74.44&31.92\\
	DenseNet-169~\cite{huang2017densely} 			& \textbf{76.39 (+0.79)}&\textbf{11.89 (4.76x)}        & 75.45 (-0.15)&13.54 (4.18x)					&75.60&56.60\\
	DenseNet-201~\cite{huang2017densely} 			& \textbf{77.06 (+0.16)}&\textbf{17.24 (4.64x)}    	   & 76.92 (+0.02)&19.09 (4.19x)					&76.90&80.06\\
      \hline
	MobileNet-v1$^*$~\cite{howard2017mobilenets}				& \textbf{70.59 (+1.02)}&4.10 (4.12x)		   		& 68.77 (-0.80)&\textbf{4.05 (4.18x)}	&69.57&16.93\\ 
	MobileNet-v2$^*$~\cite{sandler2018inverted}				& \textbf{72.19 (+0.38)}&4.25 (3.30x)		   				   & 68.06 (-3.75)&\textbf{3.45 (4.06x)}	&71.81&14.02\\
	SqueezeNet~\cite{DBLP:journals/corr/IandolaMAHDK16}				& \textbf{60.01 (+1.92)}&1.22 (1.93x)		  & 57.93 (-0.16)&\textbf{1.20 (1.96x)}	&58.09&2.35\\   
      \hline
  \end{tabular}
 \footnotesize{\\$^*$ MobileNet-v1 and MobileNet-v2 are implemented and trained by ourselves. The pre-trained models of other networks are officially provided by Pytorch.}\\
	\end{table*}

\subsection{Quantization Details}
\label{3.4}
In this section, we introduce the quantization process in details. 
Given a weight vector $\omega$ and the quantization bit $b$, the quantization process can be formulated as follow:
\begin{equation}\label{Eq3}
\hat{w} = \mathcal{L}^{-1}(Q(\mathcal{L}(w),b))
\end{equation}
where $\mathcal{L}(w)=\frac{w-\mu}{\nu}$ is a linear scaling function~\cite{DBLP:journals/corr/HeWZWYZZ16} that normalizes arbitrary vectors into the range [0,1] and $\mathcal{L}^{-1}$ is the inverse function.
Specially, as the whole parameter vector usually has a huge dimension, magnitude imbalance might push most elements in the vector to zero. This would result in an extremely harm precision. To address this issue, we adopt the bucketing technique~\cite{DBLP:journals/corr/Alistarh0TV16}, that is, the scaling function is applied separately to a fixed length of consecutive values. The length is the bucket size $k$.

In Eq.\eqref{Eq3}, $Q$ is the actual quantization function that only accepts values in [0,1]. For a certain element $w_i$ and the quantization bit $b$, this process is shown as below:

\begin{equation}\label{Eq4}
Q(w_i,b) = \frac{\lfloor w_i\,2^b\rfloor}{2^b} + \frac{\xi_i}{2^b}\\
\end{equation}
This function assigns the scaled value $w_i$ to the closest quantization point and $\xi_i$ is the rounding function as follow.
\begin{equation}
\begin{split}
 \xi_i=\left\{
\begin{aligned}
&1, \;\;\;\; \text{if}\;\; w_i\,2^b-\lfloor w_i\,2^b\rfloor > 0.5 \\
&0, \;\;\;\; \text{otherwise}\\
\end{aligned}
\right.
\end{split}
\end{equation}

Given a certain weight vector of size $N$ and the size of full precision weight $f$ (usually 32 bits), full precision requires $fN$ bits in total to store this vector. As we use $b$ bits per weight and two scaling parameter $\alpha$ and $\beta$ for each budget, the quantied vector needs $bN+ 2\frac{fN}{k}$ bits in total. Thus, the compressed ratio is $\frac{kf}{kb+2f}$ for this weight vector.

\section{Experimental Results}
In this section, we first apply our approach to existing networks and show the compression results on ImageNet. After that, we introduce the joint search results.

\begin{figure*}[ht]
  \centering
  \includegraphics[width=1.0\textwidth]{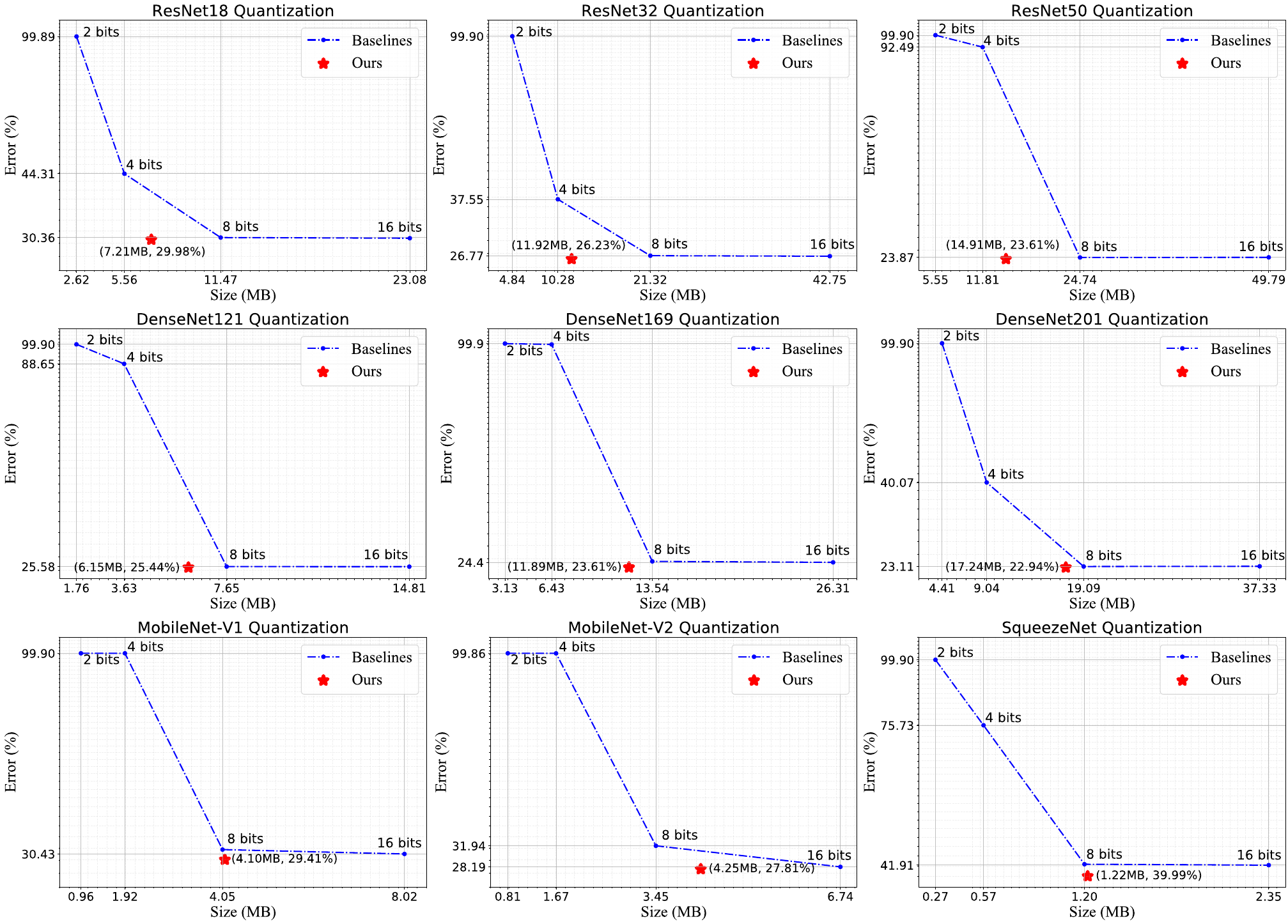}
  \caption{The results of quantization policy search for existing networks on ImageNet. Here we compare to 2bits, 4 bits, 8 bits and 16 bits models. 
 The points of Ours are clearly under the Baselines. Models quantized by our policies have better accuracies than others.}\label{fig:4} 
\end{figure*}

\subsection{Quantization on Fixed Architecture}
\label{4.1}
Our method can be flexibly applied on any existing networks for providing quantization policies.
In this section, we report the quantization results of some classical networks on ImageNet~\cite{DBLP:conf/cvpr/DengDSLL009}. These state-of-the-art networks include a series of ResNet~\cite{he2016deep}, DensenNet~\cite{huang2017densely} and some mobile size networks, e.g., MobileNet-v1~\cite{howard2017mobilenets}, MobileNet-v2~\cite{sandler2018inverted} and SqueezeNet~\cite{DBLP:journals/corr/IandolaMAHDK16}. For all ResNets~\cite{he2016deep}, DenseNets~\cite{huang2017densely} and SqueezeNet~\cite{DBLP:journals/corr/IandolaMAHDK16}, we obtain their pre-trained float models from torchvision.models class of PyTorch. Because MobileNet-v1~\cite{howard2017mobilenets} and MobileNet-v2~\cite{sandler2018inverted} models are not provided by official PyTorch, we implement and train them from scratch to get these two float models.
Table~\ref{tab:1} presents the performance of our quantization policies on the state-of-the-art networks. In the $Acc/\%$ columns, the numbers in the brackets mean the accuracy increase or decrease after compression. In the \emph{Params/M}, the numbers in the brackets mean the compression ratio. 

It is worth to note that our method can effectively improve the accuracy and compress the model size. Taking ResNet18~\cite{he2016deep} for example, the model generated by our method has 70.02\% accuracy that is 0.26\% higher than the float model. Our compressed ResNet18 has 7.21M parameters while the float model has 46.76M parameters that is 6.49 times as ours. For all these ResNets~\cite{he2016deep} and DenseNets~\cite{huang2017densely}, our method can generate models that are more accurate and smaller than both 8 bits and float models. For the mobile size networks, MobileNet-v1~\cite{howard2017mobilenets} MobileNet-v2~\cite{sandler2018inverted} and SqueezeNet~\cite{DBLP:journals/corr/IandolaMAHDK16}, ours are slightly larger than 8 bits models, but much more accurate than both the 8 bits and the float models.

In addition, we also compare our results to other compression strategies in Fig.~\ref{fig:4}, including 2 bits, 4 bits and 16 bits. It shows the bi-objective frontiers obtained by our results and the corresponding 2/4/8/16 bits results. A clear improvement appears that our results have much higher accuracies than 2/4 bits models and are much smaller than 8/16 bits models of ResNets~\cite{he2016deep} and DenseNets~\cite{huang2017densely}. For mobile size models, i.e., MobileNet-v1~\cite{howard2017mobilenets}, MobileNet-v2~\cite{sandler2018inverted} and SqueezeNet~\cite{DBLP:journals/corr/IandolaMAHDK16}, our results are more accurate than all bits models.
  
\subsection{Joint Architecture Search and Quantization}
 \label{4.2}

The joint search is conducted on CIFAR-10 to obtain the normal cell structure $\mathscr{A}_{\mathtt{nom}}$, the reduction cell structure $\mathscr{A}_{\mathtt{rec}}$ and the quantization policy $\mathscr{P}$. 
After search, we retrain CIFAR-10 and ImageNet float models from scratch. CIFAR-10 results are obtained by quantizing the float models with the search quantization policy $\mathscr{P}$.
As ImageNet architectures have additional cells and layers, it is unable to directly apply $\mathscr{P}$ on ImageNet float models. Thus we use $\mathscr{P}$ to initialize an evolution population to search ImageNet quantization policies as in Section \ref{4.1}.

In Table~\ref{tab:2}, we compare the performance of ours to other state-of-the-art methods that search only for neural architectures. Note that all methods listed in Table~\ref{tab:2} use NASNet~\cite{zoph2017learning} architecture search space.
JASQNet is obtained with $\mathcal{T_{S}}$ set as $3~\mathtt{MB}$ during search and JASQNet-Small is obtained with $\mathcal{T}_{S}$ set as $1~\mathtt{MB}$ during search. Ours-Small(float) and JASQNet (float) are the float models before the searched quantization policies applied to them.

For the model JASQNet, it achieves competitive accuracies and relatively small model size to other comparison methods. On CIFAR-10, only NASNet-A~\cite{zoph2017learning} and AmoebaNet-B~\cite{Real2018Regularized} have clearly higher accuracies than JASQNet. But their search costs are hundreds of larger times than ours. On CIFAR-10, the model size of JASQNet is more than 4 times small as the size of other comparison models. On ImageNet, the accuracy of JASQNet is competitive to others and the model size of JASQNet is also 4 times or so small as that of other comparison models.

For the model JASQNet-Small, its model size is 10 times small as the size of other comparison models on CIFAR-10. 
On ImageNet, its model size is 7 or 8 times small as others. Compared to SqueezeNet~\cite{DBLP:journals/corr/IandolaMAHDK16}, the model with similar size (41.91\% with 2.35 MB), its accuracy is much higher.

Compared to JASQNet (float) and JASQNet-Small (float), JASQNet and JASQNet-Small has a higher accuracy and smaller model size. It shows that our learned quantization policies are effective. Compared to other only searching for architecture methods, JASQNet (float) and JASQNet-Small (float) are not best. Because our search space is much larger that includes quantization choices and it is unfair to directly compare them with our float models.

  \begin{table*}[t] 
  \centering
    \caption{Comparisons to Architecture Search on CIFAR-10 and 224 ImageNet.} \label{tab:2}
  \begin{tabular}{ l  |c c| c c c | c c c} 
        \hline
           			& \multicolumn{2}{c|}{\textbf{Search Cost}} 		   & \multicolumn{3}{c|}{\textbf{CIFAR-10}} 					& \multicolumn{3}{c}{\textbf{ImageNet}}\\ 
           			
            & GPUs & Days &\#Params/M & Size/MB & Error/\% &\#Params/M & Size/MB & Error/\% \\

      \hline
      PNASNet-5~\cite{liu2017progressive}			 		& 100&1.5 					 	& 3.2 & 12.8 & 3.41 $\pm$ 0.09 & 5.1 & 20.4 & 25.8		 \\ 
      NASNet-A$^*$  ~\cite{zoph2017learning} 			& 500&4 							& 3.3	& 13.2	& 2.65					& 5.3 & 21.2&	26.0	 \\ 
      NASNet-B~\cite{zoph2017learning} 					& 500&4 							& 2.6	& 10.4	& 3.73 					& 5.3 & 21.2&	27.2	 \\ 
      NASNet-C~\cite{zoph2017learning} 					& 500&4 							& 3.1	& 12.4	& 3.59 					& 4.9 & 19.6&	27.5	 \\ 
	  AmoebaNet-B$^*$ ~\cite{Real2018Regularized} & 450&7 						& 2.8	& 11.2	& 2.55 $\pm$ 0.05	& 5.3 & 21.2& 26.0	 \\
	  ENAS$^*$ ~\cite{pham2018efficient}				& 1&0.5								& 4.6	& 18.4	& 2.89						&- &- &-	 \\ 
	  DARTS (1st order)$^*$~\cite{DBLP:journals/corr/abs-1806-09055} & 1&1.5			& 2.9	& 11.6	& 2.94				& 4.9 & 19.6 &	26.9		\\	
	  DARTS (2nd order)$^*$\cite{DBLP:journals/corr/abs-1806-09055} & 1  &4		& 3.4&13.6 &2.83$\pm$ 0.06	 &- &-&- \\
      \hline
      JASQNet (float)$^*$												& 1&3									&	3.3 &	13.2 &	2.94		 			& 4.7 & 18.8 & 27.25		\\ 
      JASQNet$^*$  															& 1&3								& 3.3 &\textbf{2.5} & 2.90		 & 4.7 & \textbf{4.9} & 27.22\\ 
      \hline
       JASQNet-Small (float)$^*$														& 1&3					& 1.8 & 7.2 & 3.08		 				&2.8 &11.2 & 34.14	\\
       JASQNet-Small$^*$  												& 1&3								& 1.8 &\textbf{0.9} & 2.97		 & 2.8 & \textbf{2.5} & 34.10\\ 
      \hline
  \end{tabular}
 \footnotesize{\\$^*$ Training with cutout~\cite{cutout} on CIFAR-10.  All methods use NASNet~\cite{zoph2017learning} architecture search space. 
 }\\
  \end{table*}
  
  \begin{figure*}[htb]
  \centering
  \includegraphics[width=0.9\textwidth]{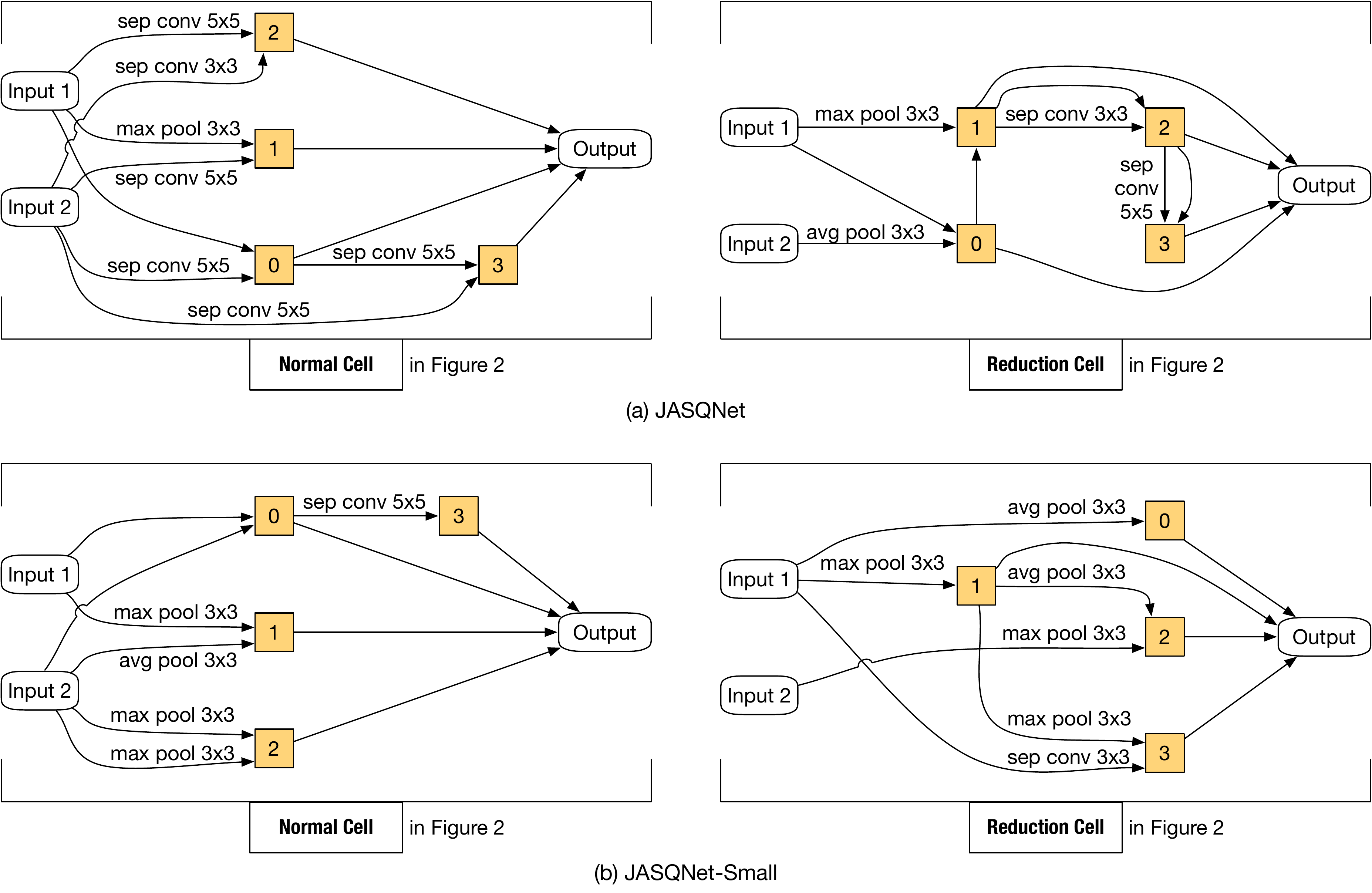}
  \caption{JASQNet and JASQNet-Small}\label{fig:5}
\end{figure*}

It is worth to clarify \#Params and Size in Table~\ref{tab:2}. \#Params means the number of free parameters and its unit is million (M). Size means model size for storage and its unit is MByte (MB). Quantization can reduce Size but not \#Params. The result architectures are shown in Fig.~\ref{fig:5}. 

  \begin{figure*}[tb]
  \centering
  \includegraphics[width=1.0\textwidth]{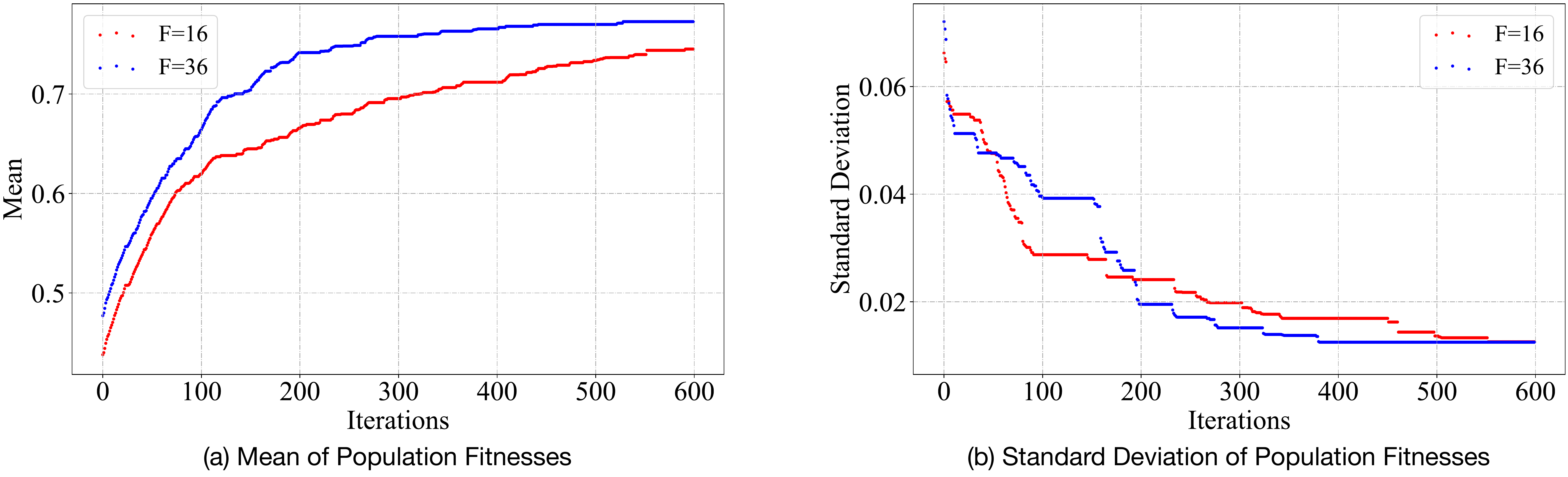}
  \caption{Ablation study on whether use small proxy networks for search.}\label{fig:6}
\end{figure*}

\subsection{Analyses}
\subsubsection{Search Process Details}
Previous works~\cite{zoph2017learning, pham2018efficient, Real2018Regularized, DBLP:journals/corr/abs-1806-09055, liu2017progressive} tend to search on small proxy networks and use wider and deeper networks in the final architecture evaluation. 
In Table~\ref{tab:3}, we list the depth and width of networks for search and networks for evaluation. N is the number of stacking cells in Fig.~\ref{fig:2} and F is the initial convolution channels.
Taking the width for example, DARTS~\cite{DBLP:journals/corr/abs-1806-09055} uses a network with initial channels 16 for search and evaluates on networks with initial channels 36. ENAS~\cite{pham2018efficient} searches on networks with initial channels 20 and evaluates on a network with initial channels 36. 

The original purpose of searching on small proxy networks is to save time. But in our joint search experiments, we empirically find it is a bit harmful to search process. We make an ablation study on using small proxy networks as in Fig.~\ref{fig:6}. The blue line represents the experiment without small proxy networks, where the networks have the same width (F=36) and depth (N=6) to those for evaluation. The red line represents searching with small proxy networks (F=16 and N=6). We keep track of the most recent population during evolution. Fig.~\ref{fig:6}~(a) shows the highest average fitness of the population over time. Fig.~\ref{fig:6}~(b) shows the lowest standard deviation of the population fitnesses over time. Wider networks might lead to higher accuracies but it is clear that the blue line in Fig.~\ref{fig:6}~(a) converges faster than the red line. Standard deviation of the population fitnesses represents the convergence of evolution. Thus, Fig.~\ref{fig:6}~(b) also shows that searching without proxy networks leads to a faster convergence.

  \begin{table}[tb] 
  \centering
    \caption{Depth and Width for Search and Evaluation on CIFAR-10.} \label{tab:3}
  \begin{tabular}{ l  |c c| c c } 
        \hline
           			& \multicolumn{2}{c|}{Search} 		   & \multicolumn{2}{c}{Evaluation}\\ 
           			
            & F & N & F & N\\

      \hline
      PNASNet-5~\cite{liu2017progressive}			 		&24 &2 					 		&48 &3  		 \\ 
      NASNet  ~\cite{zoph2017learning} 				&32 &2 							&32 	&6 		 \\ 
	  AmoebaNet ~\cite{Real2018Regularized} 		&24 &3 							&36 	&6		 \\
	  ENAS ~\cite{pham2018efficient}$^*$						&20 &2						&36 	& 5	      \\ 
	  DARTS~\cite{DBLP:journals/corr/abs-1806-09055} &16 &2					&36 	&2 	\\	
      JASQ 																		&36 &6							&36 &6  \\ 
      \hline
  \end{tabular}
   \footnotesize{\\$^*$	This info is discovered in their released code but in not their paper.}
  \end{table}

\subsubsection{Comprehensive Comparison}
\begin{figure}[tb]
  \centering
  \includegraphics[width=0.5\textwidth]{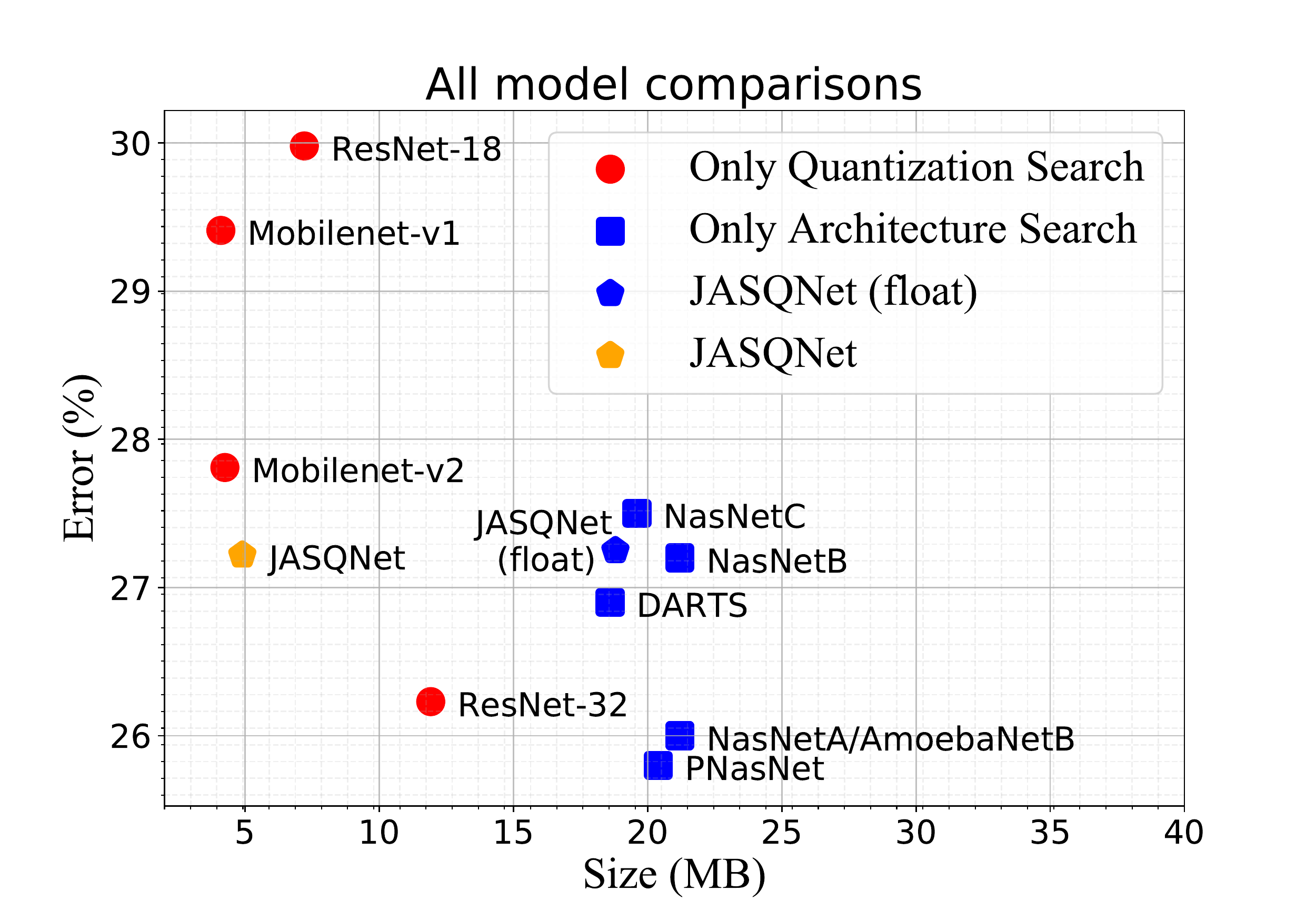}
  \caption{Comparisons with only architecture search and only quantization search. The gap between JASQNet and JASQNet (float) shows the effectiveness of our quantization policy. JASQNet reaches a better balance point thant other models.}\label{fig:7}
\end{figure}

Joint search performs better than only architecture search or only quantization search.
JASQNet are better than only architecture search (blue squares) and only quantization search (red circles) as illustrated in Fig.~\ref{fig:7}. Models with too many parameters (DenseNets), are not shown in it. It shows that JASQNet reaches a better multi-objective position. 

In addition, as shown in results in Table~\ref{tab:1}, suitable quantization policies can improve accuracy and decrease model size simultaneously. No matter for existing networks quantization or joint search of architecture and quantization, our quantized models are more accurate than their float counterparts. In Fig.~\ref{fig:7}, we also depict JASQNet (float) as a blue pentagon.
The gap between JASQNet and JASQNet (float) shows the effectiveness of our quantization policy. Their accuracies are almost same but JASQNet has much less model size.

As shown in Table~\ref{tab:2}, JASQNet (float) and JASQNet-Small (float) are not better than NASNet~\cite{zoph2017learning} or AmoebaNet~\cite{Real2018Regularized}. The first reason is that joint search results in larger search space that might harm the quality of searched architectures. The second possible reason is that their search processes spend much more computation resources than ours. 

\section{Conclusion}
Searching for both architectures and compression heuristics is a direct and convenient way for deep learning practitioners.
To our best knowledge, this task has never been proposed in the literature. 
In this work, we propose to automatically design architectures and compress models. Our method can not only conduct joint search of architectures and quantization policies, but also provide quantization policies for existing networks. The models generated by our method, JASQNet and JASQNet-Small, achieve better trade-offs between accuracy and model size than only architecture search or only quantization search.

\section*{Appendix}

\subsection*{1) CIFAR-10 Classification}

\paragraph{Dataset}
There are 50,000 training images and 10,000 test images in CIFAR-10. 5,000 images are partitioned from the training set as a validation set. 
We whiten all images with the channel mean subtraction and standard deviation division.
32 x 32 patches are cropped  from images and padded to 40 x 40. Horizontal flip is also used. We use this preprocessing procedures for both search and evaluation.

\paragraph{Training}
For fair comparisons, our training hyper-parameters on CIFAR-10 are identical to those of DARTS~\cite{DBLP:journals/corr/abs-1806-09055}. The models for evaluation are trained for 600 epochs with batch size 96 on one GPU. The version of our GPUs is Titan-XP.
The initial learning rate is 0.025 and annealed down to zero following a cosine schedule. We set the momentum rate as 0.9 and set weight decay as $3\times 10^{-4}$. Following existing works \cite{DBLP:journals/corr/abs-1806-09055,zoph2017learning,Real2018Regularized}, additional enhancements include cutout~\cite{cutout}, path dropout of probability 0.3 and auxiliary towers with weight 0.4.

\subsection*{2) ImageNet Classification}

\paragraph{Dataset}
The original input images are first resized and their shorter sides are randomly sampled in [256, 480] for scale augmentation~\cite{simonyan2014very}.  We then randomly crop images into $224 \times 224$ patches.
We also conduct horizontal flip, mean pixel subtraction and the standard color augmentation. These are standard augumentations that proposed in Alexnet~\cite{krizhevsky2012imagenet}. 
In addition,  most augmentations are excluded in the last 20 epochs with the sole exception of the crop and flip for fine-tuning.

\paragraph{Training}
Each model is trained for 200 epochs on 4 GPUs with batch size 256. 
We set the momentum rate as 0.9 and set weight decay as $4\times 10^{-5}$. We also employ an auxiliary classifier located at $\frac{2}{3} $of the maximum depth weighted by 0.4. The  initial learning rate is 0.1. It later decays with a polynomial schedule.

\subsection*{3) Quantization Process}
Previous works \cite{DBLP:journals/corr/ZhuHMD16,DBLP:journals/corr/abs-1709-01134} do not quantize the first and last layers of ImageNet models to avoid severe accuracy harm. We also follow this convention on our ImageNet models and do not apply this constraint on CIFAR-10 models.
Another detail is that we use Huffman encoding for quantized value representation to save additional space.

\subsection*{4) Search Process}
  \begin{figure}[tb]
  \centering
  \includegraphics[width=0.5\textwidth]{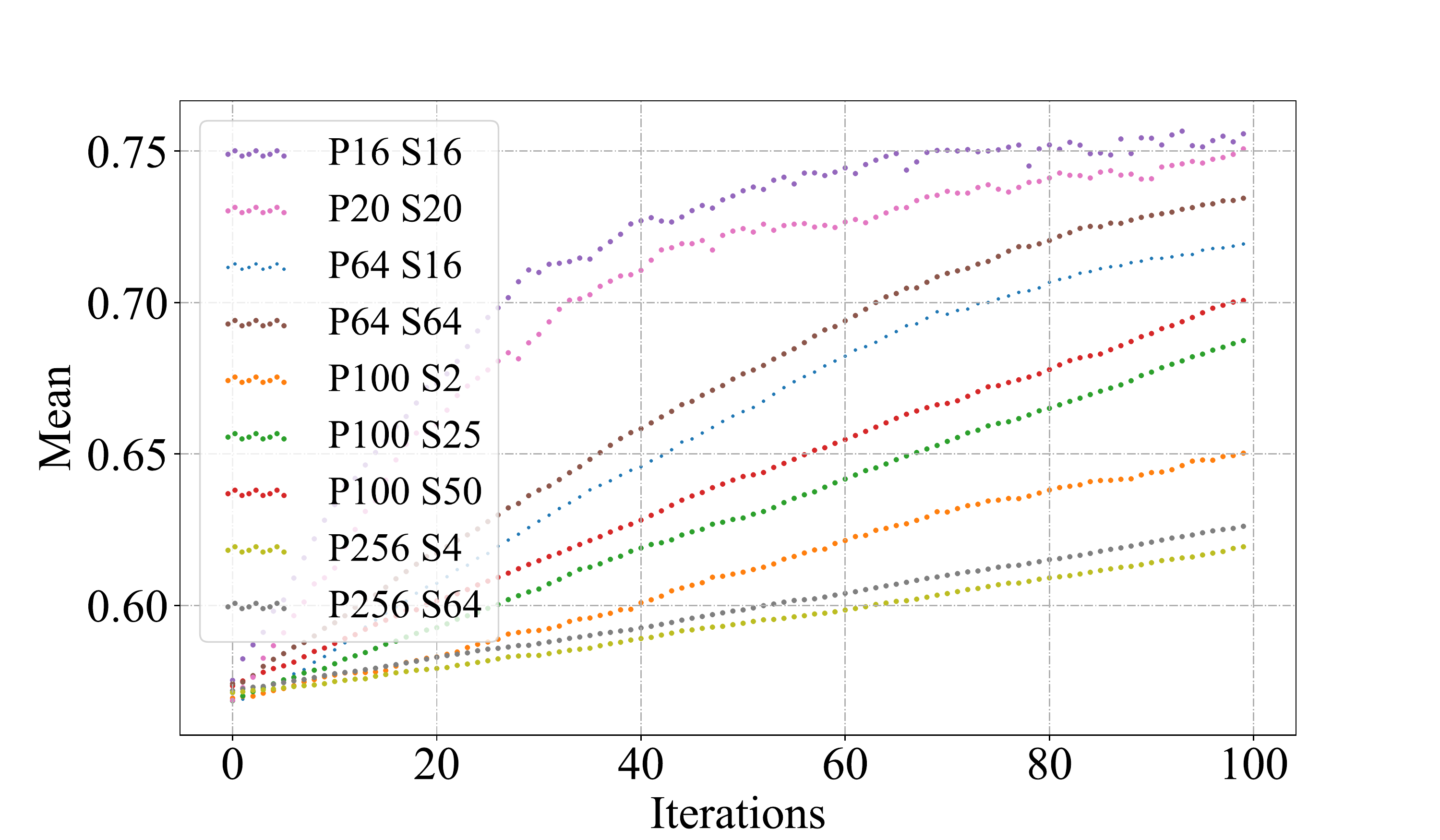}
  \caption{Hyper-parameter optimization experiments about population size and sample size. We conduct these experiments in a small scale by setting input filters F=16 and stacking cells number N=2. Each experiment runs for 100 iterations.}\label{fig:8}
\end{figure}

The evolutionary search algorithm employed in this paper can be classified into tournament selection \cite{goldberg1991comparative}. There are only two hype-parameters, population size $\mathtt{\text{\#}P}$ and sample size $\mathtt{\text{\#}S}$. 
The hyper-parameter optimization process is illustrated in Figure~\ref{fig:8}. We conduct all these experiments with the same settings except $\mathtt{\text{\#}P}$ and $\mathtt{\text{\#}S}$. 
For efficient comparison, these experiments runs in a small scale for only 100 iteration.
The input filters F is set as 16 and the stacking cells number N is set as 2. This figure shows the mean fitness of models in the population over iterations. We pick the best one ($\mathtt{\text{\#}P}=16,\mathtt{\text{\#}S}=16$) from Figure~\ref{fig:8} for the experiments in this paper.
We also employ the parameter sharing technique for acceleration \cite{pham2018efficient}, that is, a set of parameters are shared among all individual models in the population.

{\small
\bibliographystyle{ieee}
\bibliography{egbib}
}

\end{document}